\documentclass{ecai} 

\usepackage{latexsym}
\usepackage{amssymb}
\usepackage{amsmath}
\usepackage{amsthm}
\usepackage{booktabs}
\usepackage{enumitem}
\usepackage{graphicx}
\usepackage{caption}
\usepackage{multirow}
\usepackage{spverbatim}
\usepackage[ruled,vlined,linesnumbered]{algorithm2e}

\newcommand{\BibTeX}{B\kern-.05em{\sc i\kern-.025em b}\kern-.08em\TeX}

\begin{document}


\begin{frontmatter}


\paperid{123} 


\title{When to Retrieve: Teaching LLMs to Utilize Information Retrieval Effectively}


\author[A,B]{\fnms{Tiziano}~\snm{Labruna}\orcid{0000-0001-7713-7679}\thanks{Corresponding Author. Email: tlabruna@fbk.eu.}}
\author[C]{\fnms{Jon Ander}~\snm{Campos}\orcid{0000-0002-1447-5870}}
\author[D]{\fnms{Gorka}~\snm{Azkune}\orcid{0000-0002-2506-7426}} 

\address[A]{University of Bozen-Bolzano}
\address[B]{Fondazione Bruno Kessler}
\address[C]{Cohere}
\address[D]{HiTZ Center - Ixa, University of the Basque Country UPV/EHU}


\begin{abstract}
In this paper, we demonstrate how Large Language Models (LLMs) can effectively learn to use an off-the-shelf information retrieval (IR) system specifically when additional context is required to answer a given question.
Given the performance of IR systems, the optimal strategy for question answering does not always entail external information retrieval; rather, it often involves leveraging the parametric memory of the LLM itself. Prior research has identified this phenomenon in the PopQA dataset, wherein the most popular questions are effectively addressed using the LLM's parametric memory, while less popular ones require IR system usage. Following this, we propose a tailored training approach for LLMs, leveraging existing open-domain question answering datasets. Here, LLMs are trained to generate a special token, $\langle$RET$\rangle$, when they do not know the answer to a question. Our evaluation of the Adaptive Retrieval LLM (\textsc{Adapt-LLM}) on the PopQA dataset showcases improvements over the same LLM under three configurations: (i) retrieving information for all the questions, (ii) using always the parametric memory of the LLM, and (iii) using a popularity threshold to decide when to use a retriever. Through our analysis, we demonstrate that \textsc{Adapt-LLM} is able to generate the $\langle$RET$\rangle$ token when it determines that it does not know how to answer a question, indicating the need for IR, while it achieves notably high accuracy levels when it chooses to rely only on its parametric memory.  
\end{abstract}

\end{frontmatter}

\section{Introduction}


The task of question answering (QA) remains a focal point in Natural Language Understanding research. There are many different datasets serving as benchmarks for evaluating QA models, such as Natural Questions (NQ) \citep{kwiatkowski2019natural}, SQuAD \citep{rajpurkar2016squad} or QuAC \citep{choi2018quac}, just to mention a few. Nowadays, Large Language Models (LLMs) consistently outperform traditional methods on these benchmarks, showcasing remarkable performance. 

Typically, there are two primary approaches to utilize LLMs for question answering:

(i) \textbf{Closed Book Question Answering}: This approach involves strategies like instruction tuning \citep{taori2023stanford} or few-shot prompting \citep{brown2020language} to enhance performance. Here, the LLM relies solely on its parametric memory to answer questions. However, these parametric memories have inherent limitations as they are based entirely on the training corpus, meaning for example that they could be outdated regarding events occurring after the training process. 

(ii) \textbf{Open Book Question Answering}: In this approach, the LLM is coupled with an Information Retriever (IR) system \citep{izacard2021leveraging, zhu2021retrieving}. By leveraging the IR system, the LLM can retrieve relevant context to supplement its understanding and provide more accurate answers.



However, the research conducted by \citet{mallen2023not} sheds light on the complexity of question-answering strategies, challenging the notion that the optimal approach always involves the utilization of an IR system. Through the introduction of the PopQA dataset, comprising 14 thousand questions annotated with popularity scores, they demonstrated that while LLMs relying solely on their parametric memories excel in addressing high-popularity questions, the efficacy diminishes for low-popularity questions, where using IR becomes curcial. 

\begin{figure*}[t]
  \centering \includegraphics[width=0.9\textwidth]{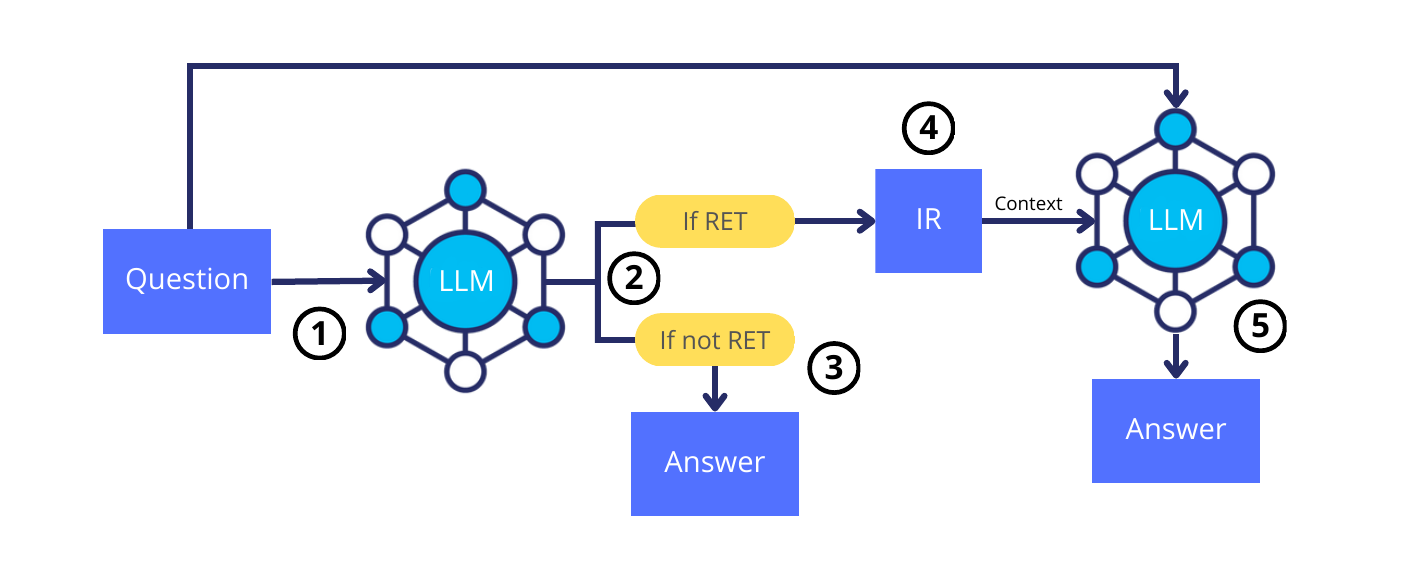}
  \caption{The inference process of \textsc{Adapt-LLM} step-by-step: given a question (step 1), an LLM decides (step 2) whether to answer the question directly (step 3) or to ask for additional contextual information, generating the special $\langle$RET$\rangle$ token; for the later, an off-the-shelf IR system is used to retrieve relevant context (step 4), which is used alongside the question to prompt again the LLM for the final answer (step 5).}
  \label{fig:process}
\end{figure*}

Their findings underscore the importance of a hybrid approach, where LLMs utilize parametric memory for high-popularity questions, but use an off-the-shelf IR system to retrieve relevant context to answer low-popularity questions. Central to their methodology is the establishment of a fixed popularity score threshold, which they use to decide whether an IR system has to be employed.

In many cases, however, question answering datasets do not include popularity scores, so relying on such scores is not a generalizable approach. Motivated by this limitation, our study aims to address whether LLMs can autonomously determine when to employ an IR system for improved question answering. To investigate this, we conduct an evaluation of an LLM using an open-domain question answering dataset to identify the questions for which the LLM provides accurate responses and those where its answers are incorrect. 

Specifically, for questions where the LLM's response is incorrect, we annotate them with a special token, $\langle$RET$\rangle$, indicating the need for additional context. Subsequently, we utilize these annotations to construct a new dataset tailored for training purposes, where we teach an LLM to answer directly if it is confident about the answer or to require context it believes is useful for answering the question (see Figure \ref{fig:process}). Our hypothesis is that through this training process, the LLM learns to use an IR system when it needs extra context to answer a question, thus we name it \textsc{Adapt-LLM}.

To validate our hypothesis, we conducted several experiments on the PopQA dataset \citep{mallen2023not}, as it provides a suitable platform for benchmarking hybrid retrieval strategies. 
As a result of these experiments we find that:

\begin{itemize}
    \item \textsc{Adapt-LLM} consistently outperforms typical fixed strategies for question answering, such as (i) using the IR system for all questions and (ii) relying solely on the parametric memory of the LLM.
    \item \textsc{Adapt-LLM} demonstrates performance comparable to strategies that rely on popularity scores to determine when to use an IR system, even without utilizing any popularity score or similar metric. It's worth noting that popularity scores are a unique feature of the PopQA dataset, rendering them inapplicable to other open-domain question answering datasets.
    \item When \textsc{Adapt-LLM} decides to retrieve additional information, the results obtained with the context are significantly better than those without it. Similarly, when \textsc{Adapt-LLM} directly answers questions relying on its parametric memory, it achieves high accuracies. These observations indicate that the model effectively discerns when to retrieve information and when it can answer a question without further context.
    \item The primary bottleneck for the performance of \textsc{Adapt-LLM} lies in the IR system. \textsc{Adapt-LLM} achieves much higher performance with gold passages compared to passages retrieved by the IR system. 
\end{itemize}

Our findings underscore the significance of adaptive retrieval strategies in enhancing the performance of LLMs for question answering tasks. By training \textsc{Adapt-LLM} to dynamically determine when to retrieve additional context, we demonstrate the feasibility of teaching an LLM how to effectively leverage external information sources only when necessary.

\section{Related Work}

Retrieval-Augmented Generation (RAG) \cite{lewis2020retrieval} has shown improvements on a wide variety of NLP areas, such as question answering \cite{karpukhin2020dense,izacard2021leveraging,seonwoo2022two,nakano2021webgpt}, truthfulness \cite{ji2023survey,lin2022truthfulqa} and language modelling \cite{guu2020realm,borgeaud2022improving,ram2023context} among others. The ability to ground model generations on retrieved text chunks has also enabled smaller models to match the performance of larger ones \cite{catav2024RAG}. Moreover, due to the extremely high cost of training LLMs, RAG has become the standard way to maintain them updated with new information, not having to re-train the models periodically to incorporate new facts \cite{gao2023retrieval}. 

Even if augmenting LLMs with retrieval is an essential step for the current generation of LLMs \cite{jiang2024mixtral,reid2024gemini} it also comes with a cost. Traditional retrieval methods as TF-IDF or BM-25 \cite{robertson2009probabilistic} are only able to retrieve documents with keyword overlap and suffer from lexical gap \cite{berger2000bridging}. In order to try to solve this issue, many pre-trained Transformer encoder based dense models have been proposed \cite{gao2021simcse,reimers2019sentence,karpukhin2020dense,gautier2022unsupervised}. Trained neural models have shown good performance over a variety of retrieval benchmarks but they still struggle in the zero-shot setup for new domains \cite{thakur2021beir}. The quality of the retrieval engine is essential for retrieval-augmented models as this will set the upper bound of the model performance. Moreover, the usage of a retrieval engine, especially when the target document index is huge, can significantly increase the latency of the model and hurt real time applications user experience \cite{barnett2024seven}. 

On the other hand, as models keep scaling, the world knowledge encoded in their parameters does too \cite{kaplan2020scaling}. Many previous efforts have shown that language models are able to memorize a significant amount of world knowledge and achieve competitive performance on tasks such as open-domain question answering when they just use their parametric knowledge for solving the task \cite{liang2023holistic,achiam2023gpt,touvron2023llama,touvron2023llama2}. 

Motivated by all this, the adaptive approach has been proposed as a new solution \cite{schick2024toolformer,mallen2023not}. In this approach, if the solution to the task is encoded in the parameters of the model, the model will be directly used for generating a solution. Conversely, if the answer is not encoded in the knowledge of the model, the answer generation will be augmented with external knowledge.

Recently, \citet{schick2024toolformer} proposed the Toolformer, a model that can self teach how and when to use external tools via simple API calls including a calculator, search engines, a calendar and so on. The self learning process is based on a synthetic text only corpus that is enriched by prompting an LLM. The LLM first adds inline API calls on top of the unsupervised corpus. These API calls are then validated by evaluating whether the execution of the API calls is helpful for predicting the future tokens. This unsupervised method significantly boosts model performance in a variety of tasks when compared against non augmented LLMs, but it also makes the model over use tools. As an example, for the QA task the model uses the search engine 99.3\% of the cases. On our work, we try to take advantage of the parametric knowledge of LLMs and just perform retrieval when needed. \textsc{Adapt-LLM} decreases the usage of IR down to 83.99\% while improving performance over vanilla retrieval.  

More similar to our work, \citet{mallen2023not} propose a dataset and method for measuring when non-parametric information needs to be retrieved. They present the PopQA dataset that contains 14K questions about a set of entities with varying popularity. The popularity of an entity is measured by the page views of its Wikipedia page. In order to solve this QA task, they use a popularity score threshold calculated on the PopQA dataset. If the popularity score of an individual entity is below the threshold they perform a retrieval step. On the contrary, if the score is greater than the threshold they directly answer the question. This method yields better results than vanilla retrieval but it requires the calculation of a popularity score that is not available in realistic QA scenarios. 

Another relevant contribution in this field, contemporaneous with our research, is the work by \citet{erbacher2024navigating}, where they trained an LLM to determine when to utilize external knowledge. They particularly focused on finding the optimal trade-off between the risk of hallucination and the cost of information retrieval, given the potentially high expense associated with IR.
Our \textsc{Adapt-LLM} method adopts a similar approach, training an LLM to learn when to retrieve information. However, we extend this by comparing our method's performance against some baselines, and assess the effectiveness of retrieving information in an adaptive manner against the strategies of never retrieving or always retrieving.\footnote{All resources are publicly available at https://github.com/tLabruna/Adapt-LLM.} 


\section{Adaptive Retrieval LLM (\textsc{Adapt-LLM})}
\label{sec:ar-llm}

Adaptive retrieval refers to the model's capability to dynamically determine whether to retrieve additional context information for generating answers in question answering tasks. Unlike traditional models that either always incorporate context or never consider it, adaptive retrieval allows the model to selectively retrieve context based on the specific requirements of each question. This adaptive approach aims to optimize performance by leveraging context only when necessary, thereby enhancing the model's ability to generate accurate answers.

As depicted in Figure \ref{fig:process}, the process of the \textsc{Adapt-LLM} unfolds in the following sequence:

\begin{enumerate}
    \item The first prompt containing the question is sent to the model (step 1 of Figure \ref{fig:process}).
    \item The \textsc{Adapt-LLM} evaluates the prompt to determine whether additional context is necessary to answer the question effectively (step 2).
    \item If the model determines that context is not required, it directly produces a response to the question by leveraging its parametric memory (step 3).
    \item If context is deemed necessary, the \textsc{Adapt-LLM} model returns a special token, represented as $\langle$RET$\rangle$, and an off-the-shelf IR system is used to retrieve pertinent context based on the question (step 4); the context is then combined with the original question prompt to form a comprehensive representation for answer generation (step 5).
\end{enumerate}

The decision-making process of \textsc{Adapt-LLM} enables the model to determine the necessity of context for answering questions through dynamic assessment of each prompt. This flexible behavior allows the model to strike a balance between utilizing context for enhanced understanding and delivering direct answers when sufficient. 

\subsection{Training \textsc{Adapt-LLM}}
\label{training}

Here, we delineate the methodology employed to train our \textsc{Adapt-LLM} model. The process of crafting the training data, denoted as $DS_{Adapt}$, is presented in Algorithm \ref{alg:training_data}.

We begin by selecting an open-domain question answering dataset containing questions $Q$, associated context passages $P$, and corresponding answers $A$. We initialize $DS_{Adapt}$ to an empty set (line 1 of the algorithm). For each question in $Q$, we leverage the base LLM without any retrieval mechanism to perform a zero-shot inference (line 3). This step allows us to differentiate questions for which the model generates correct answers from those where its responses are inaccurate. This process can be understood as a way to discover what the base LLM \textit{knows} due to its parametric memory. For questions where the model's response is accurate (line 4), we build a training set instance incorporating the following prompt, which we call \textit{parametric\_prompt}:

\begin{algorithm}[tb]
\caption{Training data creation}
\label{alg:training_data}
\KwIn{Q: questions, A: answers, P: passages, LLM} 
\KwOut{$DS_{Adapt}$: A training dataset for Adaptive Retrieval}
\SetAlgoLined
\SetKwInOut{Input}{Input}
\SetKwInOut{Output}{Output}
\BlankLine
$DS_{Adapt}$ = init\_empty()\\
\For{q, gold\_ans, pass in (Q, A, P)}{
    ans = LLM(q)

    \If{ans = gold\_ans}{
        inst = build\_instance('parametric\_prompt', q, gold\_ans)\\
        $DS_{Adapt}$.add(inst)
    }
    
    \Else{
        inst1 = build\_instance('parametric\_prompt', q, "<RET>")\\
        $DS_{Adapt}$.add(inst1)\\
        inst2 = build\_instance('context\_prompt', q, gold\_ans, pass)\\
        $DS_{Adapt}$.add(inst2)
    }
}
return $DS_{Adapt}$
\end{algorithm}


\begin{spverbatim}
Prompt: Answer the question Q. If you need help answer <RET> to get the context. Q: {...}
\end{spverbatim}
\vspace{1em}
\noindent

Alongside this prompt, we include the corresponding question from $Q$ and the golden answer from $A$, collectively forming the instance (line 5), which is subsequently appended to the $DS_{Adapt}$ dataset (line 6).


In contrast, if the LLM fails to produce a correct response to the question (line 8), we build two different instances. The first employs the same \textit{parametric\_prompt} as previously described, with $\langle$RET$\rangle$ designated as the answer (line 9), indicating the necessity for additional context. The second prompt, termed \textit{context\_prompt}, encompasses contextual information alongside the question:

\begin{spverbatim}
Prompt: Answer the question Q given the context C. Q: {...}, C: {...}
\end{spverbatim}
\vspace{1em}
\noindent

For this instance, we include the prompt, the question from $Q$, the golden answer from $A$, and the corresponding context passage from $P$ (line 11).

After populating the dataset with both types of prompts for questions where the LLM could not respond accurately and only the \textit{parametric\_prompt} with golden answers for all other questions, our training set $D_{Adapt}$ is prepared for the subsequent fine-tuning phase. The fine-tuning process entails training the base LLM on our dataset, resulting in the \textsc{Adapt-LLM} model. 

This approach ensures that the model effectively learns to discern when context is necessary for answering questions, or to provide a direct response when it suffices, as well as answer directly when provided with context.


\subsection{Inference} 
\label{inference}
In the inference phase, we utilize the fine-tuned model to generate responses to unseen questions. We employ the same prompts used during the training phase, as outlined in Section \ref{training}. 

Initially, the model is prompted to either provide a direct response or return $\langle$RET$\rangle$ if it is unsure of the answer. If the model returns $\langle$RET$\rangle$, we proceed with information retrieval to acquire relevant context by means of an off-the-shelf IR system. Subsequently, we augment the question with the retrieved context and prompt the model again using the second type of prompt introduced during the training phase.


\begin{table}[t]
\centering
\begin{tabular}{cccc}
\toprule
\textbf{Training Set} & \textbf{Model configuration} & \textbf{Accuracy} \\
\midrule
\multirow{3}{*}{\textsc{NQ}} & \textsc{Never Retrieve} & 21.43\% \\
& \textsc{Always Retrieve} & 35.86\% \\
& \textsc{Adapt-LLM} (ours) & \textbf{36.77\%} \\
\midrule
\multirow{3}{*}{\textsc{SQuAD}} & \textsc{Never Retrieve}  & 21.22\% \\
& \textsc{Always Retrieve}  & 36.59\% \\
& \textsc{Adapt-LLM} (ours) & \textbf{38.15\%} \\
\bottomrule
\end{tabular}
\caption{Performance comparison of Llama-2 models trained on the NQ and SQuAD datasets using different retrieval configurations (NR-LLM, AR-LLM, and \textsc{Adapt-LLM}), evaluated on the PopQA test set. Exact match accuracy is reported for all models.} 
\label{tab:results-exp1}
\end{table}

\section{Experiments and Results}

In this section, we outline the experimental framework aimed at assessing the performance of the proposed adaptive retrieval approach, \textsc{Adapt-LLM}. We begin by describing the datasets utilized (Section \ref{dataset}), followed by an overview of our base model (Section \ref{model}), the different configurations of the base model (Section \ref{configurations}), and the training details (Section \ref{sec:training}). Subsequently, we introduce the three primary experiments:

\begin{enumerate}
    \item Evaluation of \textsc{Adapt-LLM} performance compared to the following baseline models: (i) an LLM that retrieves contextual information for all questions, and (ii) an LLM that exclusively relies on its parametric memory without using an IR system for any question (Section \ref{sec:validating-adapt-llm}).
    \item Analysis of \textsc{Adapt-LLM}'s ability to determine when extra context is necessary to answer a question (Section \ref{sec:analysis-decision}).
    \item Comparison with the state-of-the-art approach for PopQA (Section \ref{sec:sota}).
\end{enumerate}

\begin{table}[t]
\centering
\begin{tabular}{cccc}
\toprule
 & \textbf{NQ} & \textbf{SQuAD} & \textbf{PopQA} \\
\midrule
Questions & 58,880 & 87,599 & 14,282 \\
Words/question & 9.20 & 10.06 & 6.62 \\
Words/answer & 2.26 & 3.16 & 2.04 \\
\bottomrule
\end{tabular}
\caption{Comparison of the three datasets we use for our experiments, i.e. SQuAD, NQ and PopQA. For each of them we provide the number of questions, and the average number of words per question and answer.}
\label{tab:datasets}
\end{table}

\begin{table*}[t]
\centering
\begin{tabular}{cccccc}
\toprule
\textbf{Training} & \textbf{$\langle$RET$\rangle$ Usage} & \multicolumn{2}{c}{\textbf{$\langle$RET$\rangle$}} & \multicolumn{2}{c}{\textbf{No $\langle$RET$\rangle$}} \\
& & \textbf{Acc. w/ context} & \textbf{Acc. w/o context} & \textbf{Acc. w/ context} & \textbf{Acc. w/o context} \\
\midrule
NQ & 82.26\% & 33.04\% & 14.65\% & 55.72\% & 62.36\%  \\
SQuAD & 83.93\% & 33.40\% & 9.94\% & 57.73\% & 62.92\% \\
\bottomrule
\end{tabular}
\caption{Results of the usage of the $\langle$RET$\rangle$ token in the \textsc{Adapt-LLM} model. The first column shows the percentage of PopQA questions for which the model requests additional context. The second column focuses on the questions for which \textsc{Adapt-LLM} asks for context ($\langle$RET$\rangle$), comparing the performance between answering those questions with and without context. The last column (No $\langle$RET$\rangle$) is for questions which \textsc{Adapt-LLM} decides to answer directly. We also compare the performance with and without the context retrieved by the IR system.}
\label{tab:results-exp2}
\end{table*}

\subsection{Datasets}
\label{dataset}
To ensure comprehensive training and evaluation of our models, we specifically selected three diverse question answering datasets. For training, we chose NQ \citep{kwiatkowski2019natural} and SQuAD \citep{rajpurkar2016squad}, as they are widely recognized datasets that assess factual knowledge and are based on Wikipedia. For evaluation, we opted for PopQA \citep{mallen2023not}. Below are brief descriptions of each dataset:
\paragraph{NQ} The Natural Questions dataset \citep{kwiatkowski2019natural} is a collection of real-world questions derived from Google search queries, accompanied by long-form text passages obtained from Wikipedia articles and providing a diverse range of topics and natural language variations. We utilize this dataset for \textbf{training} our models in the experiments. 
\paragraph{SQuAD} The Stanford Question Answering Dataset SQuAD \citep{rajpurkar2016squad} is a widely utilized dataset in the field of natural language processing and comprises questions posed by crowdworkers on a diverse range of Wikipedia articles, along with relevant paragraph passages serving as context. We utilize this dataset for \textbf{training} our models in the experiments. 
\paragraph{PopQA} The Popular Questions and Answers dataset \citep{mallen2023not} consists of curated questions sourced from various online platforms, encompassing a wide range of domains and styles. Given the variability in the effectiveness of context retrieval strategies observed in this dataset, we select PopQA as our test set to \textbf{evaluate} the language models' performance in determining when context is necessary for accurate answer provision.
\subsection{Base Model}
\label{model}
In our experiments, we employ Llama-2 \citep{touvron2023llama} as our base LLM. Llama-2 is an open-source instruction-based LLM, which comes in versions of 7B, 13B, and 70B parameters. The model is pretrained on an expanded corpus sourced from publicly available online data sources. This corpus offers a 40\% increase in size compared to its predecessor, contributing to the model's enhanced performance and capabilities. 

Additionally, Llama-2 features an extended context length, effectively doubling its capacity to process and comprehend longer sequences of text. These enhancements significantly improve the model's effectiveness across various natural language understanding tasks. Specifically, for our experiments, we utilize the Llama-2 model with 7B parameters, leveraging its robust capabilities for our specific research objectives.

\subsection{Model Configurations}
\label{configurations}
We conduct the experiments using three different model configurations, corresponding to the three different ways in which an LLM and an IR system can be combined:

\begin{itemize}
    \item \textbf{Adaptive Retrieval (\textsc{Adapt-LLM})}. The \textsc{Adapt-LLM} model dynamically decides whether to retrieve context based on the question and its perceived need for contextual information, as explained in Section \ref{training}. As the IR system, we use Contriever \citep{gautier2022unsupervised}, which is an unsupervised model pretrained on a large corpus, followed by fine-tuning on MS MARCO \citep{nguyen2016ms}. We only retrieve the most relevant passage according to the IR system to prompt the base LLM for the final answer. 
    \item \textbf{Never-Retrieve (NR-LLM)}. This model configuration is trained to answer questions solely based on the question text without considering any contextual information. It serves as the baseline for evaluating the performance of question answering models in the absence of context.
    \item \textbf{Always-Retrieve (AR-LLM)}. In contrast to the NR-LLM model, this configuration always retrieves context passages to assist in answering questions. It is trained to utilize context consistently for generating answers. To ensure a fair comparison with \textsc{Adapt-LLM}, we also use Contriever \citep{gautier2022unsupervised} as the IR system and only retrieve the most relevant passage as context. 
    
\end{itemize}

\subsection{Training Details}
\label{sec:training}
For all three model configurations (\textsc{Adapt-LLM}, AR-LLM and NR-LLM) and both training sets (SQuAD and NQ), we adhere to the parameter configuration established in Alpaca-Lora \citep{taori2023stanford} which includes a batch size of 128, three epochs, and a fixed learning rate of 3e-4. We incorporated LoRA (Low-Rank Adaptation) regularization, with parameters configured for r=8, alpha=16, and a dropout rate of 0.05. Training was performed on an NVIDIA A40 GPU, for an average training time of approximately 8 hours. We do not perform any model selection and we use the last checkpoint after 3 epochs of training.

\begin{figure*}[t]
  \centering
  \begin{minipage}[b]{0.45\linewidth}
    \centering
    \includegraphics[width=\linewidth]{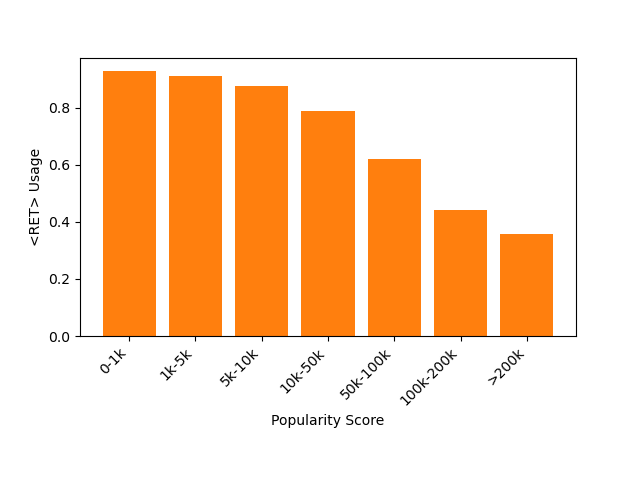}
    \label{fig:image_a}
  \end{minipage}
  \hfill
  \begin{minipage}[b]{0.45\linewidth}
    \centering
    \includegraphics[width=\linewidth]{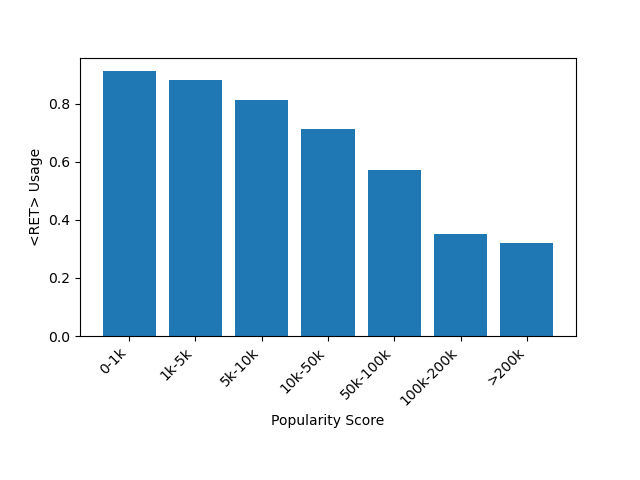}
    \label{fig:image_b}
  \end{minipage}
  \caption{Histograms depicting the proportion of questions where \textsc{Adapt-LLM} trained on NQ (left) and \textsc{Adapt-LLM} trained on SQuAD (right) ask for extra context for different popularity score intervals.}
  \label{fig:two_images}
\end{figure*}

\subsection{Validating the Adaptive Retrieval Approach}
\label{sec:validating-adapt-llm}


In order to assess the effectiveness of our adaptive approach (\textsc{Adapt-LLM}) in comparison to the NR-LLM and AR-LLM configurations, we conducted fine-tuning of the Llama-2 model on both the NQ and SQuAD datasets across all three configurations. For the NR-LLM and AR-LLM configurations, we constructed training samples by extracting question-answer pairs from the datasets and incorporating corresponding instruction prompts. 

Specifically, prompts for the NR-LLM configuration instructed the model to answer questions without additional context, whereas prompts for the AR-LLM configuration included both the question and contextual information. In contrast, the \textsc{Adapt-LLM} training set was constructed following the approach outlined in Section \ref{training}, employing a two-step process. As a result of this process, the 74.72\% of the questions in NQ are marked with the $\langle$RET$\rangle$ token, whereas the 87.49\% questions are marked for SQuAD. 


The trained models were then tested on the PopQA dataset to evaluate their performance in a real-world question answering scenario. During inference, the NR-LLM and AR-LLM models were utilized as is, with corresponding instruction prompts provided, and outputs expected to be answers to the questions. Conversely, for the \textsc{Adapt-LLM} model, we followed the same prompt procedure as explained in Section \ref{inference}. 

The generated answers are then compared to the set of possible answers for each question, which are already annotated in the PopQA test set. The evaluation metric used is \textbf{Exact Match Accuracy}, which measures the percentage of generated outputs that exactly match one of the possible answers for the corresponding question.

Table \ref{tab:results-exp1} presents the results of this experiment, illustrating the performance of the Llama-2 model across the different configurations and datasets. Across both the NQ and SQuAD training datasets, the \textsc{Adapt-LLM} configuration consistently outperforms the Never Retrieve (NR-LLM) and Always Retrieve (AR-LLM) configurations on the PopQA test set. As can be observed, NR-LLM exhibits the lowest performance among the models, with an accuracy difference of approximately 14 absolute points compared to the other configurations. This disparity suggests that the parametric memory of Llama-2 alone is not sufficient for effectively answering PopQA questions.

The differences between AR-LLM and \textsc{Adapt-LLM} are narrower. Specifically, the \textsc{Adapt-LLM} configuration achieves an accuracy of 36.77\% and 38.15\% on the PopQA test set when trained on the NQ and SQuAD datasets, respectively, compared to 35.86\% and 36.59\% for the AR-LLM configuration. Across both training datasets, \textsc{Adapt-LLM} outperforms AR-LLM, with the largest difference observed when trained on SQuAD.

All in all, these results underscore the efficacy of the adaptive retrieval approach in dynamically determining the necessity of context for accurate question answering, resulting in improved performance compared to fixed strategies of always or never retrieving context. 

Although the disparity between training \textsc{Adapt-LLM} on NQ or SQuAD is relatively minor, we try to determine the suitability of a training set for a given evaluation set. While both training sets (NQ and SQuAD) and the evaluation set (PopQA) are based on Wikipedia, subtle differences may exist.

Table \ref{tab:datasets} provides insights into the characteristics of the three datasets involved in our experimental procedure, including the total number of questions and the average number of words per question and answer. While NQ appears to be closer to PopQA in terms of question and answer lengths, the key factor influencing the better results of training \textsc{Adapt-LLM} on SQuAD may be the number of questions in the training dataset ($\sim$87K in SQuAD and $\sim$58K in NQ). Further analyses are required to elucidate the factors that render a training dataset more suitable for a given target dataset (which is beyond the scope of our study), but these results suggest that scale may play once again a crucial role.



\subsection{Contextual Retrieval Decision Analysis} 
\label{sec:analysis-decision}

In this experiment, our objective is to once again evaluate the effectiveness of the \textsc{Adapt-LLM} model, this time focusing on its ability to accurately determine when additional context is needed. 
For this purpose, we adhere to the following steps:
\begin{enumerate}
    \item We conduct inference on the \textsc{Adapt-LLM} model using the PopQA test set, prompting it to either return an answer directly or indicate the need for additional context by returning $\langle$RET$\rangle$.
    \item In the case of receiving a $\langle$RET$\rangle$ response from the \textsc{Adapt-LLM} model, we proceed with the following steps:
        \begin{enumerate}[label*=\arabic*.]
            \item We conduct inference on the \textsc{Adapt-LLM} model, prompting it to return an answer given the context obtained from the IR system.
            \item We also conduct inference on the NR-LLM model with the instruction to provide an answer directly without additional context.
        \end{enumerate}
    \item If the \textsc{Adapt-LLM} model decides to answer the question directly relying only on its parametric memory:
    \begin{enumerate}[label*=\arabic*.]
        \item We conduct inference on the \textsc{Adapt-LLM} model, prompting it to return the answer without providing context.
        \item We conduct inference on the AR-LLM model with the instruction to provide an answer using the context retrieved by the IR system.
    \end{enumerate}    
\end{enumerate}


Table \ref{tab:results-exp2} presents the results of this experiment. The first thing to note is that the \textsc{Adapt-LLM} model generates the $\langle$RET$\rangle$ token for approximately 82-83\% of the questions in the PopQA dataset, with similar ratios observed across both training datasets. This observation aligns with the low performance of the NR-LLM configuration demonstrated in Table \ref{tab:results-exp1}. 

However, \textsc{Adapt-LLM} consistently determines when additional context is required to answer a question accurately. Across both the NQ and SQuAD training datasets, \textsc{Adapt-LLM} exhibits significantly higher accuracy when retrieving context compared to the NR-LLM model's accuracy without context (as indicated in the $\langle$RET$\rangle$ column of Table \ref{tab:results-exp2}). Specifically, for the NQ dataset, the accuracy of the \textsc{Adapt-LLM} model when requesting context is 33.04\%, whereas the accuracy of the NR-LLM model without context retrieval is notably lower at 14.65\%. Similarly, for the SQuAD dataset, \textsc{Adapt-LLM} achieves an accuracy of 33.40\% with context retrieval, whereas the NR-LLM model's accuracy without context is substantially lower at 9.94\%.

\begin{table}[t]
\centering
\begin{tabular}{ccc}
\toprule
\multirow{2}{*}{\textbf{Passages}} & \textbf{SQuAD Dev} & \textbf{NQ Dev}\\
 & \textbf{Acc.} & \textbf{Acc.} \\
\midrule
Gold & \textbf{89.42\%} & \textbf{69.76\%} \\
Contriever & 22.49 & 27.04\% \\
\bottomrule
\end{tabular}
\caption{Performance comparison of \textsc{Adapt-LLM} for the SQuAD and NQ dev sets, when using the gold passages provided by the datasets and when using the best passage retrieved by Contriever.}
\label{tab:retriever}
\end{table}

Finally, the last column of Table \ref{tab:results-exp2} (No $\langle$RET$\rangle$) shows the performance of \textsc{Adapt-LLM} when answering questions based solely on its parametric memory. As can be seen, accuracies above 62\% are obtained when no context is utilized, providing further evidence that \textsc{Adapt-LLM} effectively discerns between retrieving context and providing direct answers to questions. Additionally, we evaluate the performance of these questions when context is added to the input, revealing significant decreases in accuracy of up to 7 absolute points.

These findings provide insights into the effectiveness of the decision-making process employed by the \textsc{Adapt-LLM} model in determining the necessity of additional context for accurate response generation and present empirical evidence of the necessity of performing dynamic context retrieval in improving the accuracy of question answering models.

However, it is notable that the overall performance of the model when answering questions with retrieved context, as observed in Table \ref{tab:results-exp2} (approximately 33\%), is relatively low. To further explore this observation, we conduct an additional experiment: evaluating \textsc{Adapt-LLM} (both versions trained on NQ and SQuAD) on the NQ and SQuAD development splits, comparing performance when using the gold passages of the dataset and the context retrieved by our IR system, Contriever \citep{gautier2022unsupervised}. Unfortunately, PopQA does not provide the gold passages, so direct evaluation there was not possible.

Table \ref{tab:retriever} presents the results of this experiment. A significant performance difference is observed between using the gold passage and the top passage retrieved by Contriever for both datasets (approximately 67 absolute points for SQuAD and 42 for NQ). This indicates that Contriever, and current IR systems in general, do not consistently retrieve the most relevant passage to answer a given question. This observation underscores the importance of retrieving multiple documents as context, as seen in the most successful open-domain QA systems \citep{izacard2021leveraging}, and highlights its impact on the overall performance of \textsc{Adapt-LLM} in PopQA.

To further validate the behavior of \textsc{Adapt-LLM} when requesting additional context, Figure \ref{fig:two_images} illustrates the proportion of questions for which our model generates the $\langle$RET$\rangle$ token, aggregated by popularity score intervals (left image for \textsc{Adapt-LLM} trained on NQ and right image for SQuAD). \citet{mallen2023not} suggest that high-popularity questions can be adequately answered using the parametric memory of the LLM, while lower popularity scores necessitate extra context. In Figure \ref{fig:two_images}, we observe this pattern for both versions of \textsc{Adapt-LLM}, indicating that our model, despite lacking access to popularity scores during training or inference, has learned effective criteria for requesting additional context.

\subsection{Comparison with state-of-the-art methods} 
\label{sec:sota}
We conducted a comparative analysis between our \textsc{Adapt-LLM} model and the current state-of-the-art approach for PopQA proposed by \citet{mallen2023not}. Their methodology relies on the popularity score annotated in the PopQA dataset to determine whether a question requires additional context. To establish the optimal threshold for determining question popularity, \citet{mallen2023not} split the PopQA dataset into 75\% as a development set for threshold determination and 25\% as a test set. In the original paper, they apply this methodology to various LLMs available at that moment (Llama-2 was not released yet).

To ensure a fair comparison between \textsc{Adapt-LLM} and the popularity-based method, we replicated their approach using the Llama-2 7B model to determine the best popularity score threshold (found to be 707,000) using the same PopQA development set. This allowed us to obtain results consistent with their methodology while utilizing our base LLM. Similar to the original results in \citet{mallen2023not} when using smaller models, the popularity score threshold is almost equivalent to always retrieving contextual information for Llama-2 7B. The IR usage is of 99.86\% as presented in Table \ref{tab:results-exp3}. This clearly shows how the popularity score method struggles with smaller size models, being \textsc{GPT-3 davinci-003} the only model to get a IR usage below 80\% in the original paper when using adaptive retrieval with the Contriever. Subsequently, we evaluated our \textsc{Adapt-LLM} configuration on the same 25\% test set split and compared the outcomes with those obtained using the method described by \citet{mallen2023not}. This systematic comparison enabled us to assess the efficacy of our \textsc{Adapt-LLM} model in relation to the current state of the art. 

The results of this experiment are presented in Table \ref{tab:results-exp3}. We observe comparable performance between the replicated approach of \citet{mallen2023not} and \textsc{Adapt-LLM} when trained on NQ and SQuAD datasets and tested on the 25\% subset of PopQA.  It's worth mentioning that \textsc{Adapt-LLM} does not utilize any information from PopQA, unlike \citet{mallen2023not}, who directly use the popularity score and a 75\% portion of PopQA dataset to find an optimal value for that popularity score. This methodology is not generalizable to other open-domain question answering tasks since the popularity score is a unique feature of PopQA. However, \textsc{Adapt-LLM} can be applied to any similar dataset. Given these characteristics, we believe that the results obtained by \textsc{Adapt-LLM} are even more significant, offering comparable performance to an approach that utilizes dataset-specific information. These findings substantiate the validity of our approach, demonstrating its effectiveness even when trained on datasets different from the one used for testing.

\begin{table}[t]
\centering
\begin{tabular}{ccc}
\toprule
\textbf{Model Configuration} & \textbf{IR usage} & \textbf{Accuracy} \\
\midrule
\textsc{Popularity Score} & 99.86\% & 36.81\% \\
\textsc{Adapt-LLM (NQ)} & 87.22\% & 35.30\% \\
\textsc{Adapt-LLM (SQuAD)} & 83.99\% & \textbf{37.29\%} \\

\bottomrule
\end{tabular}
\caption{Performance comparison of Llama-2 base models trained on the SQuAD and NQ datasets for the \textsc{Adapt-LLM} and \textsc{Popularity Score} configurations. The later mimics the methodology proposed by \citet{mallen2023not} with the Llama-2 LLM as the base model.}
\label{tab:results-exp3}
\end{table}

\section{Conclusions}

In this paper, we introduce \textsc{Adapt-LLM}, a LLM which learns to discern when additional context is necessary for answering a question, rather than relying solely on its parametric memory. \textsc{Adapt-LLM} is the result of fine-tuning a base LLM on an open-domain question answering dataset that has been modified to differentiate between questions answerable with the LLM's parametric memory alone and those requiring supplementary context. To construct these training datasets, we initially subject the base LLM to zero-shot evaluation to determine its accuracy in answering questions. For questions where the model's response is incorrect, we train the LLM to generate a special token, $\langle$RET$\rangle$, indicating the need for additional context.

Through extensive experiments conducted on the PopQA dataset, we show that \textsc{Adapt-LLM} performs better than its two fixed alternatives: never retrieving and always retrieving relevant context information. Furthermore, our findings highlight \textsc{Adapt-LLM}'s capability to effectively discern the necessity of additional context, which is the primary objective of this work. 

For future investigations, we propose exploring methods to enhance performance when utilizing an IR system, such as incorporating learnable sequential retrieval techniques. Furthermore, we believe it would be valuable to conduct a more in-depth analysis of the interaction between training and testing datasets in the development of \textsc{Adapt-LLM} systems.

\section{Acknowledgments}
This work received partial support from the Basque Government through research group funding IT1805-22 and the ICL4LANG project (grant no. KK-2023/00094). Additionally, we acknowledge the support of several MCIN/AEI/10.13039/501100011033 projects: (i) DeepKnowledge (PID2021-127777OB-C21) and funding from FEDER, EU; (ii) AWARE (TED2021-131617B-I00) and support from the European Union NextGenerationEU/PRTR. We express our gratitude to Carlos Domínguez for his assistance in the experimental setup and to Eneko Agirre for his valuable feedback and guidance.

\bibliography{main}

\end{document}